\documentclass{article}

\usepackage{microtype}
\usepackage{graphicx}
\usepackage{booktabs} 
\usepackage{amsmath}
\usepackage{amssymb}
\usepackage{subcaption}
\usepackage{here}

\usepackage[accepted]{icml2020}

\icmltitlerunning{Semi-supervised Anomaly Detection on Attributed Graphs}

\begin{document}

\twocolumn[
\icmltitle{Semi-supervised Anomaly Detection on Attributed Graphs}




\begin{icmlauthorlist}
\icmlauthor{Atsutoshi Kumagai}{sic,sc}
\icmlauthor{Tomoharu Iwata}{cs}
\icmlauthor{Yasuhiro Fujiwara}{cs}
\end{icmlauthorlist}

\icmlaffiliation{sic}{NTT Software Innovation Center}
\icmlaffiliation{sc}{NTT Secure Platform Laboratories}
\icmlaffiliation{cs}{NTT Communication Science Laboratories}

\icmlcorrespondingauthor{Atsutoshi Kumagai}{atsutoshi.kumagai.ht@hco.ntt.co.jp}
\icmlcorrespondingauthor{Tomoharu Iwata}{tomoharu.iwata.gy@hco.ntt.co.jp}
\icmlcorrespondingauthor{Yasuhiro Fujiwara}{yasuhiro.fujiwara.kh@hco.ntt.co.jp}

\icmlkeywords{Machine Learning, ICML}

\vskip 0.3in
]



\printAffiliationsAndNotice{\icmlEqualContribution} 

\begin{abstract}
We propose a simple yet effective method for detecting anomalous instances on an attribute graph
with label information of a small number of instances.
Although with standard anomaly detection methods it is usually assumed that instances are 
independent and identically distributed,
in many real-world applications, instances are often explicitly connected with each other,
resulting in so-called {\it attributed graphs}.
The proposed method embeds nodes (instances) on the attributed graph in the latent space
by taking into account their attributes as well as the graph structure
based on graph convolutional networks (GCNs).
To learn node embeddings specialized for anomaly detection, in which there is a class imbalance due to the rarity of anomalies,
the parameters of a GCN are trained to minimize the volume of a hypersphere that encloses 
the node embeddings of normal instances while embedding anomalous ones outside the hypersphere.
This enables us to detect anomalies by simply calculating the distances between the node embeddings and hypersphere center.
The proposed method can effectively propagate label information on a small amount of nodes to unlabeled ones
by taking into account the node's attributes, graph structure, and class imbalance.
In experiments with five real-world attributed graph datasets,
we demonstrate that the proposed method achieves better performance than various existing anomaly detection methods.
\end{abstract}

\section{Introduction}
\label{intro}
Anomaly detection is an important task in machine learning, 
which is a task of identifying anomalous instances, called anomalies, in a dataset
\cite{chandola2009anomaly,chalapathy2019deep}.
Anomaly detection methods have been used in a wide variety of applications 
such as intrusion detection
\cite{dokas2002data}, 
fraud detection
\cite{kou2004survey},
and
medical care
\cite{keller2012hics}.

Although many anomaly detection methods have been proposed such as 
one-class support vector machines (OSVM)
\cite{scholkopf2001estimating}, 
autoencoder (AE)
\cite{sakurada2014anomaly}, 
and isolation forest
\cite{liu2008isolation}, 
these methods typically assume that instances are independent and identically distributed (i.i.d.).
However, in many real-world applications, instances are often explicitly connected with each other, i.e., they have graph structures.
For example, in botnet detection on the Internet, each host is connected by its communication
\cite{bilge2012disclosure}.
In anomalous user detection on social networking services,
users are connected by their social relationships
\cite{egele2015towards}.
Such graphs, i.e., each node in a graph has instance (attributes) information,
are called {\it attributed graphs} or {\it attributed networks}.

To detect anomalies on attributed graphs,
many methods have been proposed
\cite{li2017radar,peng2018anomalous,ding2019deep,li2019specae,akoglu2015graph}.
By considering graph  structure as well as instance information,
these methods often perform better than anomaly detection methods for i.i.d. data.
Most existing methods aim to find anomalies on attributed graphs in an unsupervised fashion, i.e., 
not considering the label (normal and anomalous) information of nodes.
However, label information, which is valuable for anomaly detection,
may be usable in practice.
Semi-supervised learning methods for an attributed graph can use this label information to classify unlabeled instances
\cite{kipf2016semi,yang2016revisiting,hamilton2017inductive,wu2019simplifying,rong2019truly,rong2019truly2,sen2008collective}.
Although these methods are effective for standard classification tasks,
they do not perform well when the number of anomalous training instances is small,
which is common in anomaly detection tasks due to their rarity.

\begin{figure}[t]
\centering
\includegraphics[width=8.3cm]{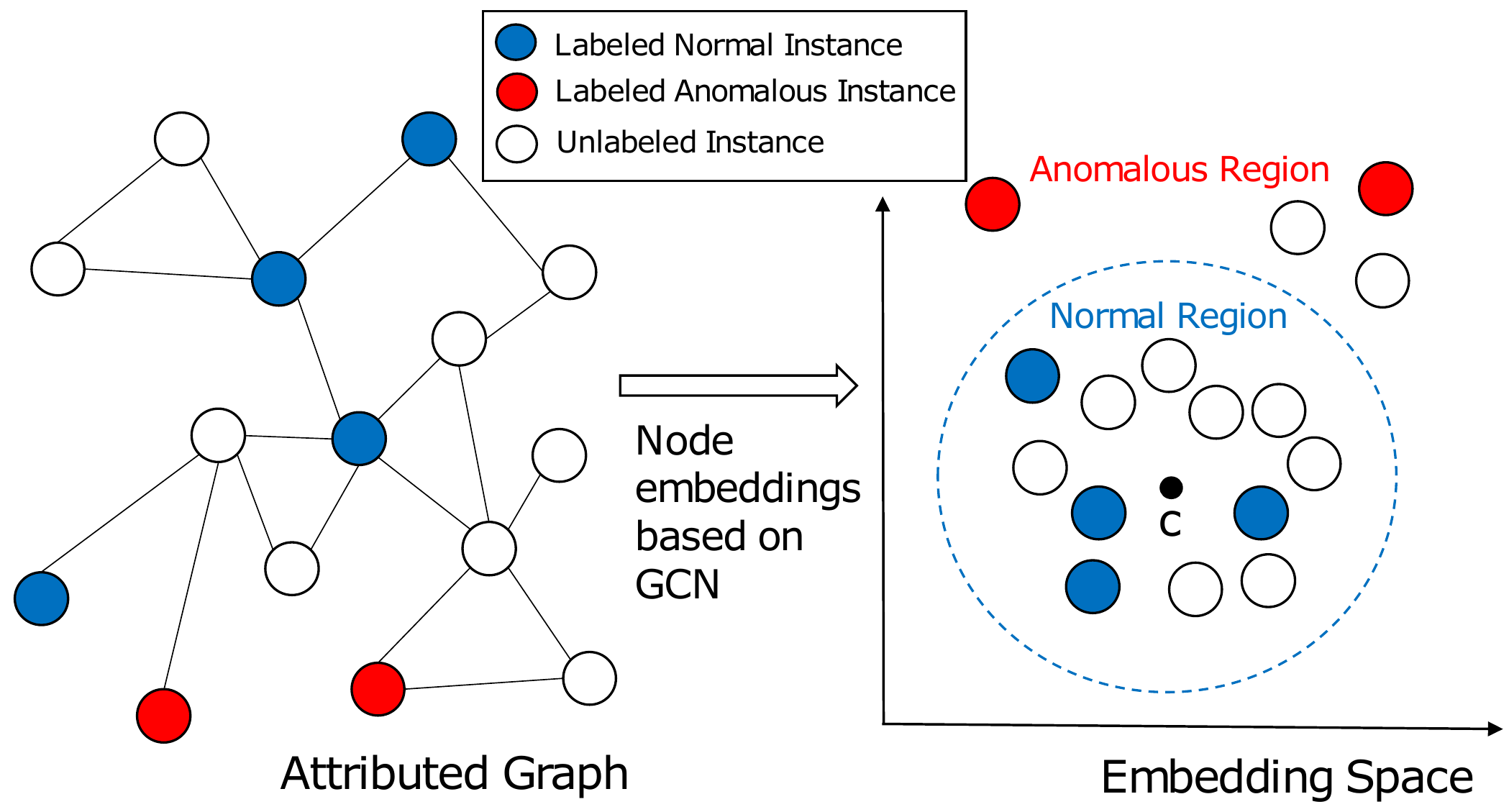}
\caption{Overview of the proposed method. The proposed method learns the parameters of the GCN so that 
the node embeddings of normal instances are enclosed into a hypersphere characterized by center vector ${\bf c}$,
and those of anomalous instances are far away from the center vector in the embedding space.
The proposed method learns the node embeddings while considering the node's attributes,
graph structure, and class imbalance.
After learning, the proposed method detects anomalies on the basis of the distance between the instances and the center vector.}
\label{overview}
\end{figure}

In this paper, we propose a novel semi-supervised anomaly detection method for an attribute graph
in which there is a class imbalance.
We focus on detecting anomalies on the attributed graph
by using the graph structure as well as labeled and unlabeled instance information
\footnote{Although the term ``semi-supervised'' sometimes means using normal instances only for training in the anomaly detection literature,
we use it to mean using both labeled and unlabeled instances following a previous study
\cite{ruff2019deep}.}.
In the proposed approach, we embed all nodes on the attribute graph in the latent space to better discriminate
between anomalous and normal instances.
Specifically, the proposed method is based on graph convolutional networks (GCNs)
\cite{kipf2016semi},
which can output node embeddings given an attributed graph while considering both graph structure and instance information effectively by stacking graph convolutional layers.
With the proposed method, to obtain node embeddings specialized for anomaly detection, 
the parameters of the GCN are trained to minimize the volume of a hypersphere that encloses the node embeddings of normal instances while embedding anomalous ones outside the hypersphere.
In anomaly detection tasks, it is important to model a data description of the normal class because the number of anomalies is too small for their description to be modeled
\cite{chandola2009anomaly,chalapathy2019deep}.
By minimizing the volume of the hypersphere enclosing normal node embeddings, 
we can effectively learn a brief data description of the normal class, which is effective
in one-class classification tasks
\cite{tax2004support,ruff2018deep}.
In addition, to embed anomalous instances outside the hypersphere,
the proposed method uses a differential area under the curve (AUC) loss as the regularizer,
which can effectively extract anomalous information even though the number of anomalous training instances is small
\cite{iwata2019supervised,kumagai2019transfer}.
Even if a small amount of nodes have label information on the attributed graph,
this information can be effectively propagated to other nodes by using both the graph structure and the attributes of 
all nodes with the GCNs.
As a result, the proposed method can accurately detect anomalies on the attributed graph.
Figure \ref{overview} illustrates the proposed method.

Our main contributions are summarized as follows:
\begin{itemize}
\item We proposed a simple yet effective semi-supervised anomaly detection method on attributed graphs. 
Our method learns node embeddings specialized for anomaly detection in such a way that
normal node embeddings are placed in a hypersphere while anomalous ones lie outside the hypersphere
based on GCNs.
\item Through the experiments using five real-world attributed graph datasets, we demonstrated that the proposed method performs better than various existing anomaly detection methods.
\end{itemize}

\section{Related Work}
\label{related}
Anomaly detection, which is also called outlier detection or novelty detection, has been widely studied
\cite{chandola2009anomaly,chalapathy2019deep}.
Many unsupervised anomaly detection methods have been proposed such as OSVM based methods
\cite{scholkopf2001estimating,chalapathy2018anomaly},
AE based methods
\cite{sakurada2014anomaly,zhou2017anomaly,chen2017outlier},
and density based methods
\cite{zong2018deep,mahadevan2010anomaly}.
Recently, deep one-class classification
\cite{ruff2018deep}
which is closely related to the proposed method,
has showed promising results on i.i.d. data.
This method learns compact representations of instances by minimizing a data-enclosing hypersphere in output space,
which is an extension of the classical support vector data description
\cite{tax2004support}
to the deep learning.
However, this method is not applicable to the attributed graph.
The proposed method utilizes this data-enclosing hypersphere approach to learn node embeddings for the normal instances on the attributed graph.
Some studies focus on supervised anomaly detection methods that use both anomalous and normal information to obtain
anomaly detectors
\cite{iwata2019supervised,chawla2002smote,yamanaka2019,ruff2019deep}.
All these methods assume that instances are i.i.d. and cannot use any graph structure
information.

Unsupervised anomaly detection methods on attributed graphs have been proposed
\cite{akoglu2015graph}.
Early methods use graph structure information only such as node degree, common neighbors, and egonet features to detect anomalies on the graph
\cite{xu2007scan,akoglu2010oddball,tong2011non}.
Recent methods
such as residual analysis based methods
\cite{li2017radar,peng2018anomalous}
and graph AE based methods
\cite{ding2019deep,li2019specae}
use node instance information as well as graph structure information to improve performance.
These methods do not use label (anomalous and normal) information.
In contrast, the proposed method uses the label information to improve anomaly detection performance.

Semi-supervised learning for graph structured data has been proposed
\cite{zhu2003semi,zhou2004learning,sen2008collective}.
These methods propagate label information of a small amount of nodes to unlabeled nodes
by using both node instances and a graph structure.
By taking advantage of the progress of graph neural networks (GNNs) including GCNs,
these methods have achieved state-of-the art results on various semi-supervised node classification tasks
\cite{kipf2016semi,yang2016revisiting,hamilton2017inductive,wu2019simplifying,rong2019truly,rong2019truly2}.
However, these methods do not assume the class imbalance and 
thus are not appropriate for anomaly detection tasks.

Few methods aim to detect anomalies on attributed graphs considering both label information and class imbalance like the proposed method.
ImVerde
\cite{wu2018imverde} 
is a semi-supervised learning method based on random walks considering the class imbalance.
One method for rare category characterization
\cite{zhou2018sparc}
uses curriculum learning to learn node representations.
In addition to learn classifier networks,
both methods use the random-walks for learning node embeddings, 
which have many hyperparameters such as walk lengths, context sizes, and sampling numbers
\cite{perozzi2014deepwalk,grover2016node2vec,tang2015line}.
The proposed method is based on simple GCNs for classifiers and thus will be convenient in practice.
In addition, these methods are difficult to apply to one-class classification tasks, which contain only normal training instances.
In contrast, the proposed method can be used since it explicitly models representations of the normal class.

\section{Proposed Method}
\label{method}

\subsection{Task}
Let ${\cal G} = ({\cal V}, {\bf A}, {\bf X})$ be an undirected attributed graph, 
where ${\cal V}$ represents a set of nodes $\{v_1, \dots, v_{N} \}$, 
${\bf A} \in \mathbb{R}^{N \times N}$ represents a symmetric adjacency matrix where
its $(n,m)$-th element $a_{nm} > 0$ denotes that there is an edge between node $v_n$ and $v_m$ and $a_{nm}=0$ denotes a no-edge, 
and ${\bf X} = [ {\bf x} _1, \dots, {\bf x} _N ]^{\top}  \in \mathbb{R}^{N \times D}$ represents
the set of instances, where ${\bf x} _n$ is $D$-dimensional attribute vector of $n$-th node $v_n$. 
The index sets for anomalous and normal training nodes are represented as 
${\cal A} = \{ n | v_n {\rm \ is \ anomalous} \}$ and $\ {\cal N} = \{ n | v_n {\rm \ is \ normal} \}$, respectively.
We assume that the label information is given for a small amount of nodes on the attribute graph, i.e., $|\mathcal{A} \cup \mathcal{N}| \ll N$.
In addition, we assume the class imbalance, i.e., $|{\cal A}|  \ll |{\cal N}|$, since anomalies rarely occur.
We note that the proposed method is applicable even when there are no anomalous nodes, $|{\cal A}| = 0$.

Our task is to estimate anomaly scores of unlabeled nodes on the graph, ${\cal V} \setminus ( {\cal A} \cup {\cal N} )$, 
so that the anomaly score becomes high (low) when the instance is anomalous (normal),
given the attributed graph ${\cal G}$ and its label information ${\cal A} \cup {\cal N}$. 

\subsection{Anomaly Scores}
We define the anomaly score for each node as follows:
\begin{align}
a(v_n) := \Vert {\bf h} _n - {\bf c} \Vert^2,
\end{align}
where ${\bf h} _n \in \mathbb{R}^{K}$ is a $K$- dimensional learned node embedding for the $n$-th node,
${\bf c} \in \mathbb{R}^{K} $ is a pre-determined center vector,
and $\Vert \cdot \Vert$ is the Euclidian norm.
This anomaly score takes a small (large) value when node embedding ${\bf h} _n$ is close to (far from) center vector ${\bf c}$.
Therefore, to detect anomalies accurately, we want to learn node embeddings in such a way
that node embeddings for normal instances are placed close to center vector ${\bf c}$ while
those of anomalous instances are far away from ${\bf c}$.
We will explain how to learn these node embeddings in the next subsection.

\subsection{Model}
The proposed method learns node embeddings specialized for anomaly detection on the attributed graph 
on the basis of GCNs, which are proposed by Kipf and Welling
\yrcite{kipf2016semi}.

The GCNs learn $K$-dimensional node embedding ${\bf h} _n \in \mathbb{R}^{K}$ for the $n$-th node
by applying multiple layer transformations to attribute vector ${\bf x} _n$.
Specifically, the $(\ell+1)$-th layer ${\bf H}^{(\ell +1)} = [ {\bf h}_1^{(\ell+1)}, \dots, {\bf h}_{N}^{(\ell+1)} ]^{\top} $ is calculated from the previous $\ell$-th layer ${\bf H}^{(\ell)}$with the following propagation rule:
\begin{equation}
\label{gcn}
{\bf H}^{(\ell +1)} = \sigma ({\bf \tilde{{D}}}^{-\frac{1}{2}} {\bf \tilde{A} } {\bf \tilde{{D}}}^{-\frac{1}{2}} {\bf H}^{(\ell)} {\bf W} ^{(\ell)} ),
\end{equation}
where ${\bf \tilde{A} } = {\bf A} + {\bf I}$ is the adjacency matrix of the graph ${\cal G}$ with added self-connections,
${\bf \tilde{{D}}} \in \mathbb{R} ^{N \times N} $ is the degree matrix of ${\bf \tilde{A} }$, which is a diagonal matrix where its $(n,n)$-th element is $\sum_{m=1}^{N} a_{nm} + 1$, ${\bf W} ^{(\ell)} $ is a layer-specific trainable
weight matrix, and $\sigma(\cdot)$ is an activation function such as the ReLU.
The initial node embeddings are set to the original attribute vectors,
${\bf H}^{(0)} := {\bf X}$.
Note that the proposed method is applicable even when the graph does not have attributes ${\bf X}$ by regarding the identity matrix as ${\bf X} = {\bf I} $ as described in previous studies
\cite{mehta2019stochastic}.
By decomposing equation \eqref{gcn} for each node,
we can see that ${\bf h} _n^{(\ell+1)} $ is represented as follows:
\begin{align}
{\bf h} _n^{(\ell+1)} &= \sigma \left( \frac{1}{d_n + 1} {\bf W}^{{(\ell)}^{\top}} {\bf h} _n^{(\ell)} \right. \nonumber\\
& \left. + \sum _{m=1}^{N} \frac{a_{nm}}{\sqrt{(d_n + 1)(d_m + 1)} } {\bf W}^{{(\ell)}^{\top}} {\bf h} _m^{(\ell)} \right),
\end{align}
where $d_n$ is the degree of the $n$-th node, $d_n:=\sum_{m=1}^{N} a_{nm}$.
This equation means that node embeddings of the next layer are calculated using node embeddings of its connected nodes and 
the node itself at the current layer.  
By applying $L$ layer transformations, information on nodes within the $L$-hop neighborhood can be used to learn the node embedding.
Therefore, the GCNs can learn node embeddings while considering attribute information of all nodes with their
graph structure by stacking multiple layers.
The $L$-th layer is regarded as the node embeddings of the proposed method, ${\bf H} = [ {\bf h}_1, \dots, {\bf h}_{N} ]^{\top}  := {\bf H}^{(L)}$.
The propagation rule \eqref{gcn} can be understood as a first-order approximation of localized spectral
filters on graphs.
For the details, please refer to the paper
\cite{kipf2016semi}.
Although we used the GCNs as building blocks of our model since
they are simple and thus will be convenient in practice,
we can use any graph neural networks to
obtain node embeddings,  such as GraphSAGE
\cite{hamilton2017inductive}
and GAT
\cite{velivckovic2017graph}.

We construct the objective function of the proposed method that consists of two terms.
We first explain the first term of the objective function.
To learn the data description of the normal class,
the proposed method minimizes the volume of the hypersphere that encloses the node embeddings of 
normal instances.
Specifically, the first term of objective function with normal instances to be minimized is as follows:
\begin{align}
{\cal L} _{{\rm nor}} (\theta) := \frac{1}{|{\cal N}|} \sum_{n \in {\cal N}} a(v_n) = \frac{1}{|{\cal N}|} \sum_{n \in {\cal N}} \Vert {\bf h} _n - {\bf c} \Vert^2,
\end{align}
where ${\bf \theta}$ is set of trainable weight matrixes of the GCN $\theta = \{ {\bf W}^{(0)}, \dots, {\bf W}^{(L-1)} \}$.
Minimizing the mean squared Euclidian distance of the normal node embeddings 
to hypersphere center ${\bf c}$ forces the GCN to learn the compact normal data representations.

Second, we explain the second term of the objective function.
To use anomalous instance information effectively,
we employ a differential approximation of the AUC,
which is effective for class imbalance data
\cite{herschtal2004optimising,iwata2019supervised,kumagai2019transfer}:
\begin{align}
\label{auc}
{\cal R} _{{\rm AUC}} (\theta) := \frac{1}{|{\cal A}| |{\cal N}|} \sum _{n \in {\cal A}} \sum _{m \in {\cal N}}
f(a(v_n)-a(v_m)),
\end{align}
where $f(\cdot)$ is sigmoid function; $f(x) = \frac{1}{1+{\rm exp}(-x)}$.
Since $f(\cdot)$ takes the maximum value one when $a(v_n) \gg a(v_m)$ and the minimal value zero 
when  $a(v_n) \ll a(v_m)$, maximizing \eqref{auc} encourages the score of anomalous instances to be higher than 
those of normal ones even though the number of anomalous instances is small.

The final objective function of the proposed method to be minimized is a weighted sum of ${\cal L} _{{\rm nor}} (\theta)$
and ${\cal R} _{{\rm AUC}} (\theta)$:
\begin{align}
\label{final}
{\cal L} (\theta) := {\cal L} _{{\rm nor}} (\theta) - \lambda {\cal R} _{{\rm AUC}}(\theta),
\end{align}
where $\lambda \ge 0$ is the hyperparameter that controls the influence of the differentiable AUC loss.
By omitting the AUC regularizer or $\lambda=0$, 
this objective function can be used even though there is no anomalous label information.
By adding the AUC regularizer, the proposed method can learn more sophisticated node embeddings
that can detect anomalies accurately.
Note that, when there are no anomalous instances or $\lambda=0$, 
we should use the GCNs without bias terms, use unbounded activation functions such as the ReLU, and avoid 
using an all-zero vector as center vector ${\bf c}$ to prevent a hypersphere collapse, in which any node embedding converge to the center vector ${\bf c}$ \cite{ruff2018deep}.

The parameters of the GCN can be optimized by minimizing \eqref{final} with any gradient-based optimization methods.
Even though few nodes have label information, 
the proposed method can effectively and efficiently propagate this information to other nodes 
on the basis of the GCN.

\subsection{Estimation}
After learning the parameters of the GCN by minimizing \eqref{final},
the anomaly score of unlabeled instance on the graph $a(v_{\ast})$ is obtained as $a(v_{\ast}) = \Vert {\bf h} _{\ast} - {\bf c} \Vert^2$.
Although the proposed method described in this paper is transductive, which means we can only estimate
anomaly scores of instances that are already observed in the graph at training time,
we can easily extend it to inductive, which means we can estimate anomaly scores for unobserved instances at training time, by applying inductive variants of GNNs such as Planetoid
\cite{yang2016revisiting} 
and GraphSage
\cite{hamilton2017inductive}. 

\section{Experiments}
We demonstrated the effectiveness of the proposed method using five real-world attributed graph datasets.
To measure anomaly detection performance on the attributed graphs,
we used AUC, which is a well used measure for anomaly detection tasks.
All experiments were conducted on a Linux server with an Intel Xeon CPU and a NVIDIA GeForce GTX 1080 GPU.

\subsection{Data}
We used five real-world attributed graph datasets: Cora, Citeseer (Cite), Pubmed (Pub), Amazon-Photo (Photo), and Amazon-Computers (Comp).

Cora, Cite, and Pub are public datasets widely used in previous studies
\cite{kipf2016semi,yang2016revisiting,wu2018imverde,wu2019simplifying}\footnote{https://github.com/kimiyoung/planetoid}.
All of them are citation networks, each node corresponds to one scientific publication,
and the edge represents the citation relationship between two publications.
Each publication is represented by a bag-of-words attribute vector.
Photo and Comp are also well used public datasets
\cite{shchur2018pitfalls}\footnote{https://pytorch-geometric.readthedocs.io/en/latest/}.
These datasets are segments of the Amazon co-purchase graph
\cite{mcauley2015image},
where nodes represent goods and the edge indicates that two goods are frequently bought together.
Each product review is represented by a bag-of-words attribute vector.
For all datasets, the edges are unweighted, i.e., $a_{nm}=1$ if there is a link between $v_n$ and $v_m$.
Each attribute was linearly rescaled to $[0,1]$.
The statistics of the datasets are summarized in Table \ref{table:dataset}.
Although these datasets have several classes, 
we created a binary class problem for each dataset by regarding the smallest class as anomalous
and the remaining classes as normal following the previous studies
\cite{wu2018imverde,zhou2018sparc}.
The average anomaly rates of Cora, Cite, Pub, Photo, and Comp are 0.06, 0.07, 0.21, 0.04, and 0.02, 
respectively.
For each dataset, we evaluated anomaly detection performance by changing the ratio of labeled instances and all instances within $\{2.5\%, 5\%, 10\% \}$.
For each case, we used 10\% of all instances for validation and the remaining for testing instances.
We randomly generated ten training/validation/testing datasets for each case
and evaluated the average test AUC over ten sets. 

\begin{table}
\centering
\caption{The statistics of datasets.}
\label{table:dataset}
\begin{tabular}{lrrrr} \hline
Data & Nodes & Edges & Attributes & Classes \\ 
\hline
Cora & 2,708 & 5,278 & 1,433 & 7 \\
Cite & 3,327 & 4,732 & 3,703 & 6 \\
Pub & 19,717 & 44,338 & 500 & 3 \\
Photo & 7,487 & 119,043 & 745 & 8 \\
Comp & 13,381 & 245,778 & 767 & 10 \\ 
\hline\end{tabular}
\end{table}

\subsection{Comparison Methods}
We evaluated two variants of the proposed method: Ours-AN and Ours-N.
Ours-AN is the method explained in Section \ref{method}, which uses both anomalous and normal label information.
Ours-N uses only normal label information.
The proposed method was implemented by using PyTorch
\cite{paszke2017automatic}
and PyTorch Geometric
\cite{fey2019fast}.

We compared the proposed method with the following nine methods.

{\bf OSVM} is the one-class support vector machine
\cite{scholkopf2001estimating}.
The OSVM finds the maximal margin hyperplane which separates the given normal data from
the origin in a RKHS. We used the RBF kernel.

{\bf DOC-N} is the deep one-class classification (One-class Deep SVDD)
\cite{ruff2018deep}, 
which is a recently proposed unsupervised anomaly detection method for i.i.d. data.
This method uses the feed-forward neural network to output embeddings and aims to minimize the volume of the hypersphere that encloses the embeddings of normal instances.
Although this objective is also used in the proposed method, DOC-N cannot use any graph structure information.

{\bf DOC-AN} is a supervised extension of DOC-N.
We added the differentiable AUC regularizer to the objective function of the OCD-N.
We included this method in the comparison methods to evaluate the effectiveness of considering the attributed graph structure 
in the proposed method.

{\bf DSAD} is the deep semi-supervised anomaly detection
\cite{ruff2019deep},
which is an extension of DOC-N for semi-supervised anomaly detection. This method uses anomalous and normal labeled instances and unlabeled instances to learn the anomaly detector for i.i.d. data.

{\bf NN} is the feed-forward neural network classifier for i.i.d. data.
The parameters of NN are trained by minimizing the cross entropy loss.

{\bf SLGCN} is the semi-supervised learning method based on the GCNs
\cite{kipf2016semi}.
The parameters of the GCNs are trained by minimizing the cross entropy loss of labeled nodes.

{\bf DW} is the DeepWalk
\cite{perozzi2014deepwalk},
which is a famous unsupervised node embedding method for graph structured data.
This method learns node embeddings on the basis of skip gram models
\cite{mikolov2013efficient},
which are applied to the sequences of random-walks.
We used negative sampling instead of hierarchical softmax to improve performance the same as
\cite{grover2016node2vec}.
After learning node embeddings, logistic regression was used as classifiers the same as
\cite{wu2018imverde,zhou2018sparc}.

{\bf DOM} is the Dominant
\cite{ding2019deep},
which is a recently proposed unsupervised anomaly detection method on attributed graphs.
This method uses an autoencoder framework to reconstruct the original attributed graph (graph structure and node 
attributes). 
The anomaly scores are defined as a weighted sum of the graph structure and node attribute reconstruction errors.

{\bf ImVerde} is a recently proposed semi-supervised anomaly detection method for class imbalanced attributed graphs
\cite{wu2018imverde}.
This method learns node embeddings on the basis of a vertex-diminished random-walk model that
reduces the transition probability to one node each that it has visited to deal with the class imbalance.
We used the authors' implementation\footnote{https://github.com/jwu4sml/ImVerde}.

OSVM and DOC-N are anomaly detection methods for i.i.d. data, which learn from normal training instances.
Note that DOC-N uses unlabeled instances as well as normal instances for training assuming that
there are fewer anomalies than normal instances in the original paper.
However, training with only normal instances showed better results in our experiments,
so we report the results of DOC-N learned with normal instances in this paper.
NN and DOC-AN are supervised learning methods for i.i.d. data, 
which learn from both anomalous and normal training instances.
DSAD learns from anomalous and normal training instances and unlabeled instances.
We used DSAD in the transductive setting. i.e., unlabeled training instances are equivalent to testing instances, for a fair comparison.
These five methods do not use any graph structure information.
DOM uses both graph structure and instance information but not label information.
DW uses both the graph structure and anomalous and normal label information.
SLGCN and ImVerde use the graph structure, instances, and anomalous and normal label information,
which is the same as Ours-AN.
For the proposed method, DOC-N, DOC-AN, DSAD, NN, and SLGCN,
the three-layer feed-forward neural networks with $32$ hidden nodes
and ReLU activation were used.
For DOM, the three-(two-)layer feed-forward neural network with $32$ hidden nodes and
ReLU activation was used for the encoder (the decoder).
For ImVerde, we used the three-layer feed-forward neural network for the classifier the same as the authors' implementation.
\begin{table*}[t]
\caption{Average and standard deviation of test AUCs [\%] when 2.5\% of all instances are labeled.}
\label{table:auc1}
\centering
\scalebox{0.84}{
\begin{tabular}{lccccccccccc}
\hline
Data & \multicolumn{1}{c}{Ours-AN} &  \multicolumn{1}{c}{Ours-N} & \multicolumn{1}{c}{OSVM} & \multicolumn{1}{c}{DOC-N} & 
\multicolumn{1}{c}{DOC-AN} & \multicolumn{1}{c}{DSAD} & \multicolumn{1}{c}{NN} & \multicolumn{1}{c}{SLGCN} & \multicolumn{1}{c}{DW} &
\multicolumn{1}{c}{DOM} & \multicolumn{1}{c}{ImVerde} \\
\hline
Cora & \textbf{88.8(5.4)} & 62.6(9.9)  & 50.0(0.1)  & 57.7(3.9)  & 72.7(6.2) & 69.6(6.5)  & 72.7(6.0)  & 84.9(6.9) & 52.1(4.0) & 52.3(0.9) & \textbf{85.9(6.1)}  \\
Cite & \textbf{65.6(4.7)}  & 56.0(4.4) & 50.6(0.4) & 55.3(1.6) & 59.9(6.5) & 53.8(2.9) & \textbf{63.1(5.0)}  & 60.9(6.0)  & 49.4(2.4) & 53.9(0.6) & 60.3(6.5)  \\
Pub  & 95.6(0.3) & 76.5(4.2) & 68.9(0.9)  & 73.7(6.3) & 95.1(0.4) & 91.3(2.4) & 95.3(0.3)  & \textbf{96.2(0.1)} & 52.8(1.3)  & 50.8(0.4) & 94.3(0.5)  \\
Photo  & \textbf{95.4(1.8)}  & 55.1(11.) & 51.9(0.6) & 52.3(1.4)  & 84.0(5.8) & 81.9(5.7) & 87.1(3.9) & 90.1(2.5) & 64.1(5.6)  & 38.1(0.4) & 89.1(1.4)  \\
Comp & \textbf{98.8(0.3)} & 56.9(5.1) & 47.3(0.7) & 46.6(1.5)  & 94.1(2.2)  & 92.2(2.5) & 95.8(1.6) & 98.1(0.3) & 87.0(4.9)  & 46.8(1.2)  & \textbf{98.5(0.7)}  \\
\hline
Avg & \textbf{88.9(13.)} & 61.4(11.) & 53.7(7.8) & 57.1(10.)  & 81.2(14.) & 77.8(15.) & 82.8(14.) & 86.0(14.) & 61.1(15.) & 48.4(5.8) & 85.6(14.) \\
\hline
\end{tabular}
}
\end{table*}
\begin{table*}[t]
\caption{Average and standard deviation of test AUCs [\%] when 5\% of all instances are labeled.}
\label{table:auc2}
\centering
\scalebox{0.84}{
\begin{tabular}{lccccccccccc}
\hline
Data & \multicolumn{1}{c}{Ours-AN} &  \multicolumn{1}{c}{Ours-N} & \multicolumn{1}{c}{OSVM} & \multicolumn{1}{c}{DOC-N} & 
\multicolumn{1}{c}{DOC-AN} & \multicolumn{1}{c}{DSAD} & \multicolumn{1}{c}{NN} & \multicolumn{1}{c}{SLGCN} & \multicolumn{1}{c}{DW} &
\multicolumn{1}{c}{DOM} & \multicolumn{1}{c}{ImVerde} \\
\hline
Cora & \textbf{91.8(5.4)} & 67.1(5.8) & 50.2(0.1)  & 58.3(2.8)  & 74.6(6.2) & 72.4(5.7) & 77.2(4.1)  & 89.2(7.6)  & 56.2(4.5) & 52.5(0.9) & \textbf{91.1(3.2)} \\
Cite & \textbf{68.3(3.8)} & 57.4(3.1) & 50.7(0.5)  & 56.0(0.8) & 62.3(3.5)  & 61.1(4.2) & \textbf{66.6(2.2)}  & 63.9(4.7)  & 50.7(2.4) & 53.9(0.7) & 64.5(4.5)  \\
Pub & 96.2(0.2) & 76.1(4.8) & 71.0(1.1)  & 79.9(3.4) & 96.0(0.3) & 91.7(1.7) & 96.1(0.2) & \textbf{96.6(0.1)}  & 54.1(1.3) & 53.0(0.5) &  94.9(0.5) \\
Photo & \textbf{97.0(0.7)} & 56.2(9.6) & 51.9(0.6)  & 52.3(0.8) & 90.1(2.4) & 89.8(3.1) & 91.8(1.6) &  92.3(1.2)  & 71.0(1.5) & 38.1(0.5) & 92.2(1.2) \\
Comp & \textbf{99.1(0.3)} & 58.2(5.8) & 47.2(0,8) & 47.0(1.4)  & 95.6(1.5)  & 92.8(2.6) & 97.1(0.7) & 98.3(0.3) & 90.5(1.8) & 47.2(1.2) & 98.6(0.6)  \\
\hline
Avg & \textbf{90.5(12.)} & 63.0(9.2) & 54.2(8.7) & 58.7(12.)  & 83.7(14.) & 81.6(13.) & 85.8(12.) & 88.1(13.) & 64.5(16.) & 48.9(6.0)  & 88.3(13.) \\
\hline
\end{tabular}
}
\end{table*}
\begin{table*}[!]
\caption{Average and standard deviation of test AUCs [\%] when 10\% of all instances are labeled.
}
\label{table:auc3}
\centering
\scalebox{0.84}{
\begin{tabular}{lccccccccccc}
\hline
Data & \multicolumn{1}{c}{Ours-AN} &  \multicolumn{1}{c}{Ours-N} & \multicolumn{1}{c}{OSVM} & \multicolumn{1}{c}{DOC-N} & 
\multicolumn{1}{c}{DOC-AN} & \multicolumn{1}{c}{DSAD} & \multicolumn{1}{c}{NN} & \multicolumn{1}{c}{SLGCN} & \multicolumn{1}{c}{DW} &
\multicolumn{1}{c}{DOM} & \multicolumn{1}{c}{ImVerde} \\
\hline
Cora & \textbf{95.4(2.7)} & 72.3(7.0) & 51.8(1.7) & 59.8(4.8) & 82.1(3.4) & 72.9(3.3)  & 83.5(3.0) & \textbf{94.5(4.3)} & 56.3(6.2) & 52.6(0.9) & \textbf{94.5(2.1)} \\
Cite & \textbf{72.9(4.3)} & 60.1(2.1) & 51.1(0.9) & 56.7(1.7) & 67.0(3.7) & 62.4(4.4) & 68.7(4.0) & 68.6(3.3) & 51.4(2.4) & 53.9(0.8) & 68.5(6.9) \\
Pub & \textbf{96.6(0.1)} & 73.4(5.7) & 73.1(0.8) & 76.3(2.8) & \textbf{96.7(0.2)} & 93.2(1.0) & \textbf{96.7(0.1)} & \textbf{96.7(0.1)} & 54.6(1.1) & 53.1(0.4) & 95.5(0.4)  \\
Photo & \textbf{97.9(0.3)} & 53.6(3.9)  & 51.7(0.9) & 53.1(0.6) & 92.5(1.4) & 89.5(1.9) & 93.8(1.0) & 93.3(0.8) & 72.7(1.3) & 38.1(0.7) & 92.8(1.0) \\
Comp & \textbf{99.1(0.3)} & 58.5(4.6) & 47.4(0.8) & 47.5(1.0)  & 96.8(0.7) & 93.5(1.9) & 97.2(0.5) & 98.3(0.3) & 91.8(1.1) & 46.6(1.9) & \textbf{99.1(0.4)} \\
\hline
Avg & \textbf{92.4(10.)} & 63.6(9.4) & 55.0(0.8) & 58.7(10.) & 87.1(12.) & 82.3(13.) & 88.0(11.) & 90.3(11.) & 65.4(15.) & 48.8(6.1)  & 90.1(13.)  \\
\hline
\end{tabular}
}
\end{table*}
\begin{figure*}[t]
  \begin{minipage}{0.33\hsize}
      \centering
      \includegraphics[width=6.0cm]{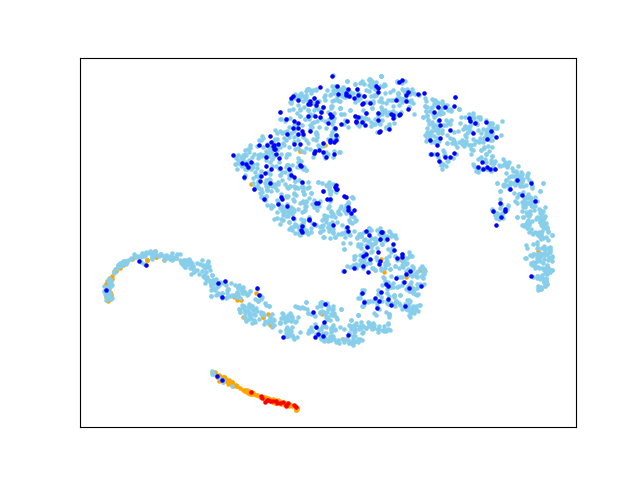}
      \subcaption{Ours-AN}
  \end{minipage}
  \begin{minipage}{0.33\hsize}
      \centering
      \includegraphics[width=6.0cm]{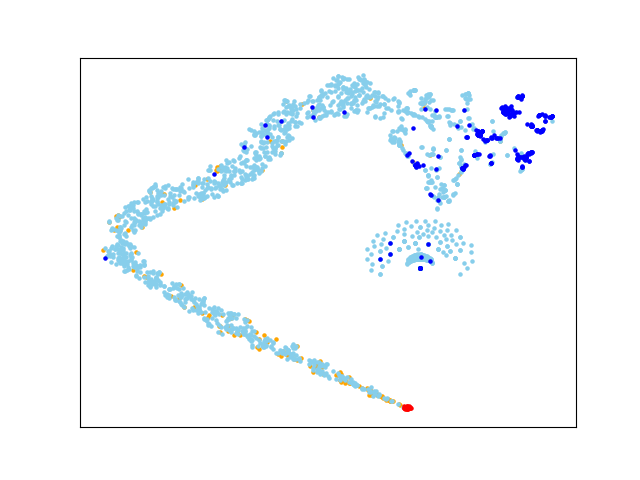}
      \subcaption{DOC-AN}
  \end{minipage}
   \begin{minipage}{0.33\hsize}
      \centering
      \includegraphics[width=6.0cm]{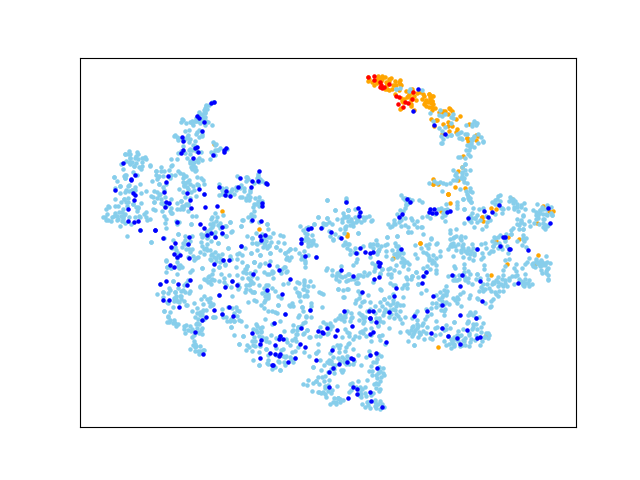}
      \subcaption{SLGCN}
  \end{minipage}
  \begin{minipage}{0.33\hsize}
      \centering
      \includegraphics[width=6.0cm]{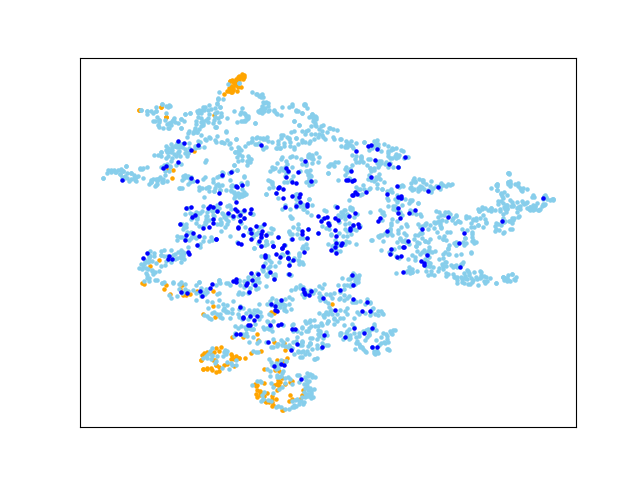}
      \subcaption{Ours-N}
  \end{minipage}
  \begin{minipage}{0.33\hsize}
      \centering
      \includegraphics[width=6.0cm]{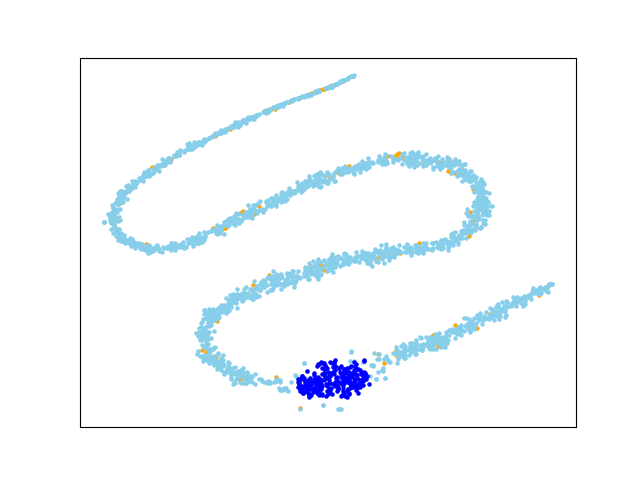}
      \subcaption{DOC-N}
  \end{minipage}
   \begin{minipage}{0.33\hsize}
      \centering
      \includegraphics[width=6.0cm]{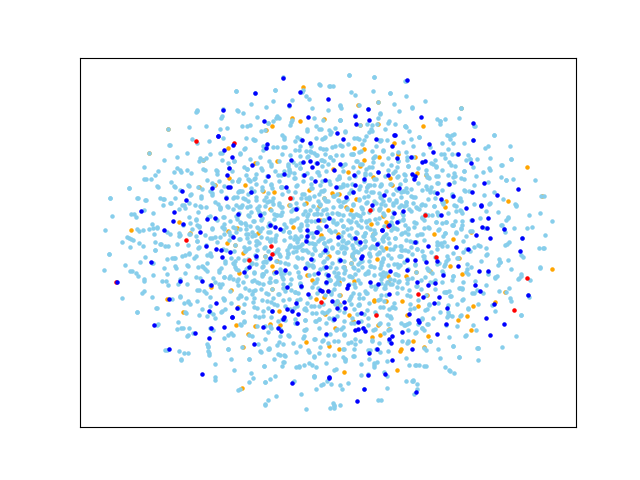}
      \subcaption{DW}
  \end{minipage}
  \caption{Visualization of the learned node embeddings on Cora. Red and blue points represent anomalous and normal training instances, respectively. Orange and sky blue points represent anomalous and normal testing instances, respectively.}
  \label{fig_emb_an}
\end{figure*}

\subsection{Hyperparameters}
For Ours-AN, DOC-AN, DSAD, NN, SLGCN, DW, and ImVerde, we selected hyper-parameters by using validation AUC.
For Ours-N, OSVM, DOC-N, and DOM, 
hyper-parameters were selected on the basis of the average anomaly score on validation normal instances 
since these methods do not use any anomalous label information for training.
For OSVM, the kernel parameter was selected from $\{ 10^{-3},10^{-2},\dots,10^3 \}$.
For DW, 
we used the following typical parameters:
the length of a walk was $80$, 
the window size of neighbor in a random walk sequence was $10$,
and
the number of negative samples was $10$.
For logistic regression of DW, regularization parameter $C$ was chosen from $\{ 10^{-2},10^{-1},\dots,10^{2}\}$.
For DOM, the balancing parameter of structure and attribute reconstruction $\alpha$
was set to 0.5, which is the recommended value in the original paper.
For ImVerde, we followed the parameters used in the author's implementation.
For the proposed method, DOC-N, DOC-AN, DSAD, DW, DOM, and ImVerde,
the dimension for node embeddings $K$ was set to $32$.
For Ours-AN and DOC-AN, regularization parameter for the AUC regularizer $\lambda$
was chosen from $\{1,10,\dots,10^4\}$.
For DSAD, the weighting parameter for labeled instances $\eta$ was selected from $\{10^{-2}, 10^{-1}, \dots, 10^{4}\}$.
For DOC-N and DSAD, we set weight regularization parameter as $10^{-6}$ the same as in the original papers.
In addition, we used the weights from the encoder part of trained AE for initialization
and set hypersphere center ${\bf c}$ to the mean of the node embeddings for normal instances after
performing an initial forward pass, which is a recommended procedure
\cite{ruff2018deep,ruff2019deep}. 
Following this, for Ours-N, we used the same procedure except for changing the AE as the graph AE
\cite{kipf2016variational}.
For Ours-AN and DOC-AN,
we did not use pre-training weights since both methods worked well without pre-training weights due to the AUC
regularizer in our preliminary experiments.
We set hypersphere center ${\bf c}$ to the mean of the node embeddings for normal instances after performing an initial forward pass.
For all methods, we used the Adam optimizer
\cite{kingma2014adam}
with a learning rate of $0.001$, and
the maximum number of epochs was 500, 500, 500, 1000, and 1000 for Cora, Cite, Pub, Photo, and Comp, respectively.
We used early-stopping based on the validation data to avoid over-fitting. 

\subsection{Results}
First, we quantitatively evaluated the anomaly detection performance of the proposed method.
Tables \ref{table:auc1} -- \ref{table:auc3} show the average and standard deviation of test AUCs for each dataset
with the different ratio of labeled and all instances.
In Tables \ref{table:auc1} -- \ref{table:auc3}, boldface denotes the best and comparable methods according
to the paired t-test at a significance level of 5\%.
Ours-AN showed the best/comparable test AUCs in almost all cases (13 of 15).
Overall, methods that use anomalous label information (i.e., Ours-AN, DOC-AN, DSAD, NN, SLGCN, and ImVerde)
performed better than other methods, 
which indicates the usefulness of anomalous label information for anomaly detection tasks.
Although DW uses both anomalous and normal label information, it performed poorly.
This was most likely because this method takes a two-step approach, i.e., learning classifiers after learning node embeddings, 
and thus discriminative information disappeared in the process of learning node embeddings. 
As for methods that do not use anomalous label information (i.e., Ours-N, OSVM, DOC-N, and DOM),
Ours-N performed the best in almost all cases (13 of 15).
Although DOC-N, DOC-AN, and DSAD aim to minimize the volume of the instances-enclosing hyperspheres like the proposed method, the proposed method outperformed them 
because it takes the graph structure information into account.
As a result, these results indicate the effectiveness of the proposed method in settings in which
both anomalous and normal labels or only normal labels are available for training. 

Next, we visualized the learned node embeddings
to quantitatively evaluate the proposed method.
Figure \ref{fig_emb_an} shows the node embeddings on Cora learned by Ours-AN, Ours-N, DOC-N, DOC-AN, DW, and SLGCN
when 10\% of all instances were labeled.
For SLGCN, we used the hidden layer as the node embeddings.
We used t-distributed stochastic neighbor embeddings (t-SNE)
\cite{maaten2008visualizing}
to reduce the dimensions of the node embeddings
from $32$ to 2.
As for methods that use both anomalous and normal label information for learning node embeddings
(i.e., Ours-AN, DOC-AN, and SLGCN), Ours-AN was able to learn better node embeddings than the others in that 
it separated normal and anomalous node embeddings well.
Although DOC-AN separated anomalous and normal training instances well,
it did not generalize to testing instances because it does not consider the attributed graph structures.
Ours-AN was able to learn useful node embeddings for detecting anomalies
by using the attributed graph structures.
As for methods that do not use anomalous label information for learning node embeddings, i.e., Ours-N, DOC-N, and DW, Ours-N was able to learn better node embeddings than the others in that anomalous instances was
located at the end of region of normal node embeddings.
Like DOC-AN, DOC-N did not generalize to testing instances although it embedded normal training instances into the small volume region.
DW did not learn good node embeddings because it does not use any label information to learn node embeddings.
Overall, these results showed that the proposed method can learn useful node embeddings for anomaly detection tasks on the attribute graph.
\begin{figure*}[t]
  \begin{minipage}{0.195\hsize}
      \centering
      \includegraphics[width=3.55cm]{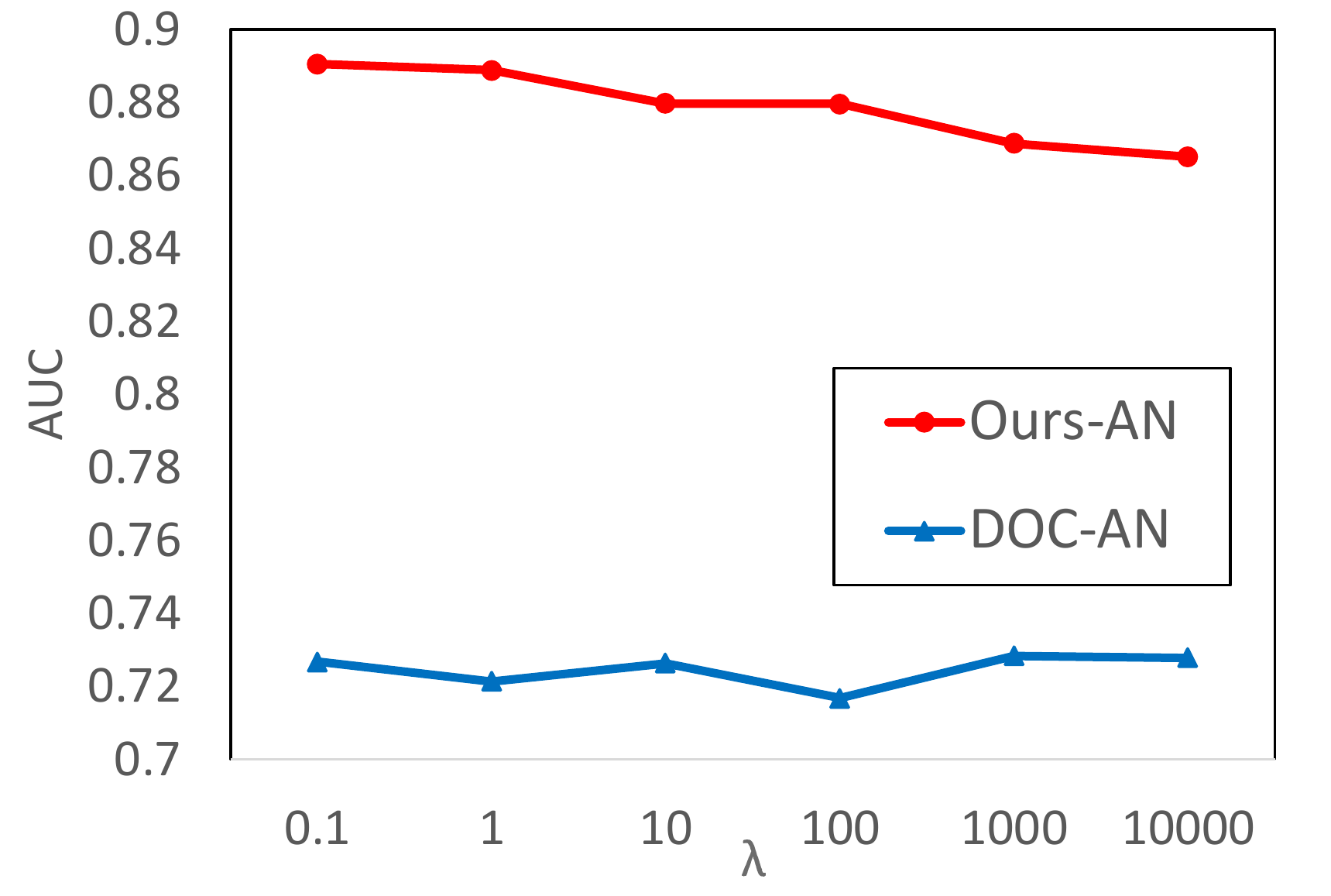}
      \subcaption{Cora}
  \end{minipage}
  \begin{minipage}{0.195\hsize}
      \centering
      \includegraphics[width=3.55cm]{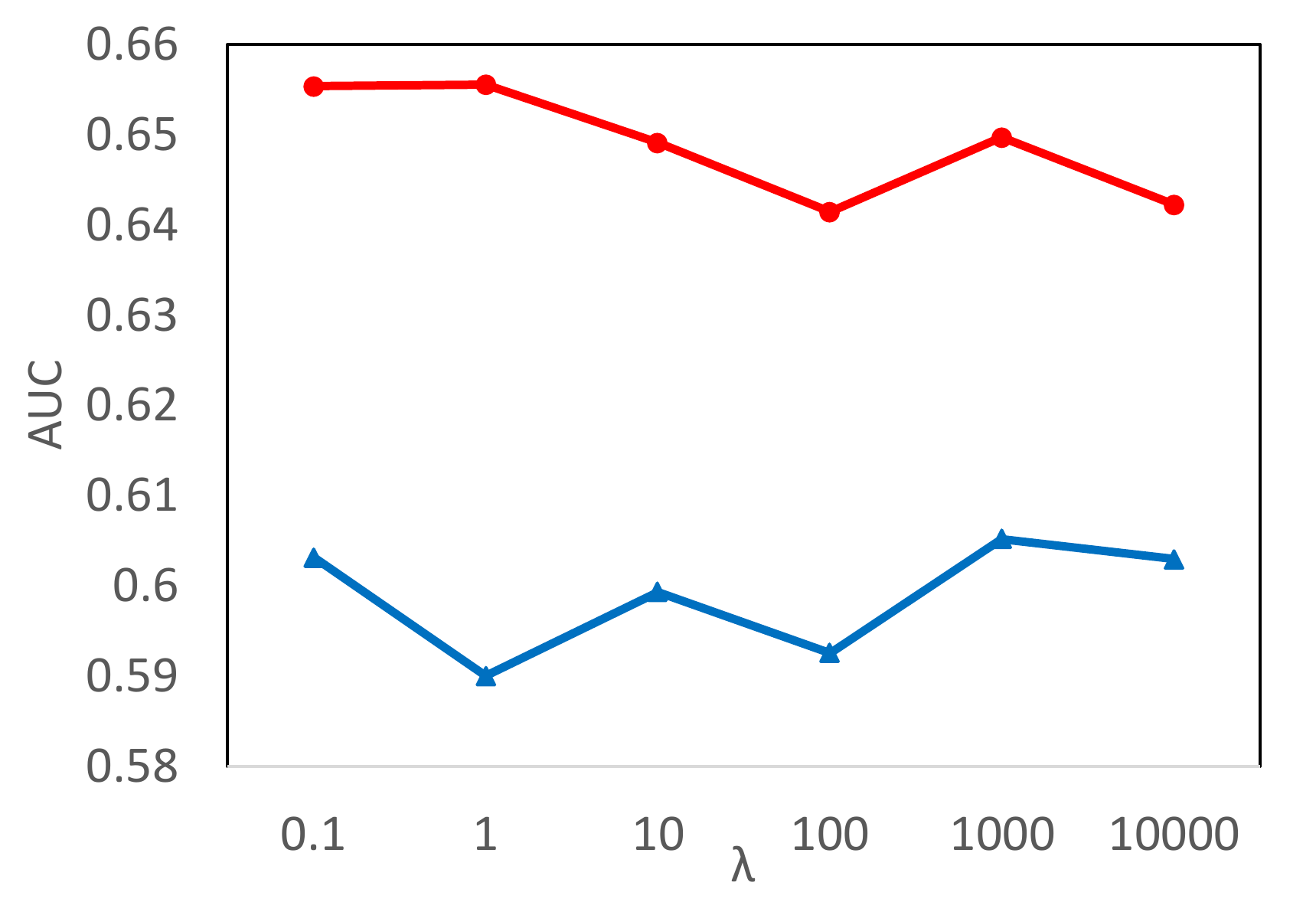}
      \subcaption{Cite}
  \end{minipage}
  \begin{minipage}{0.195\hsize}
      \centering
      \includegraphics[width=3.55cm]{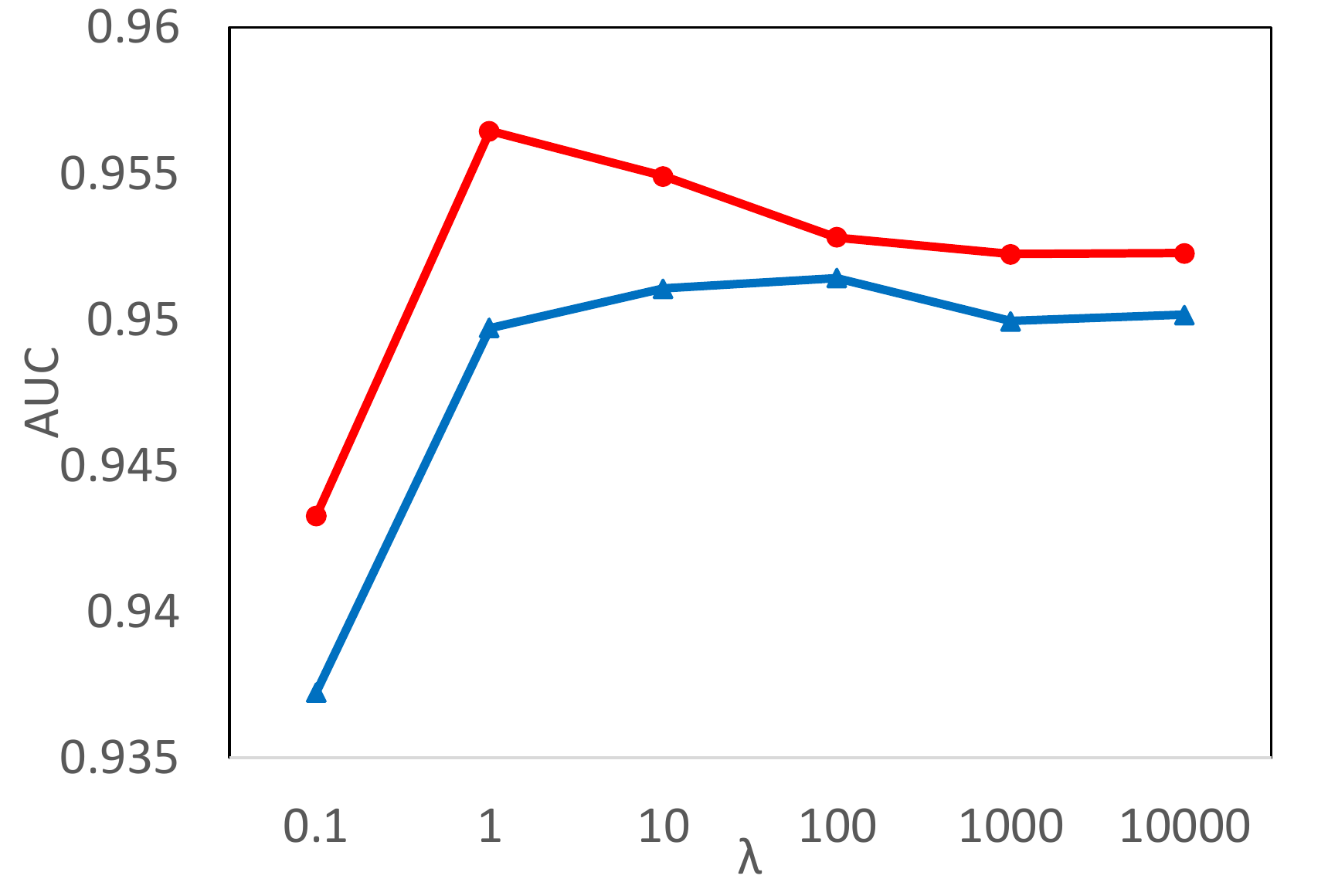}
      \subcaption{Pub}
  \end{minipage}
   \begin{minipage}{0.195\hsize}
      \centering
      \includegraphics[width=3.55cm]{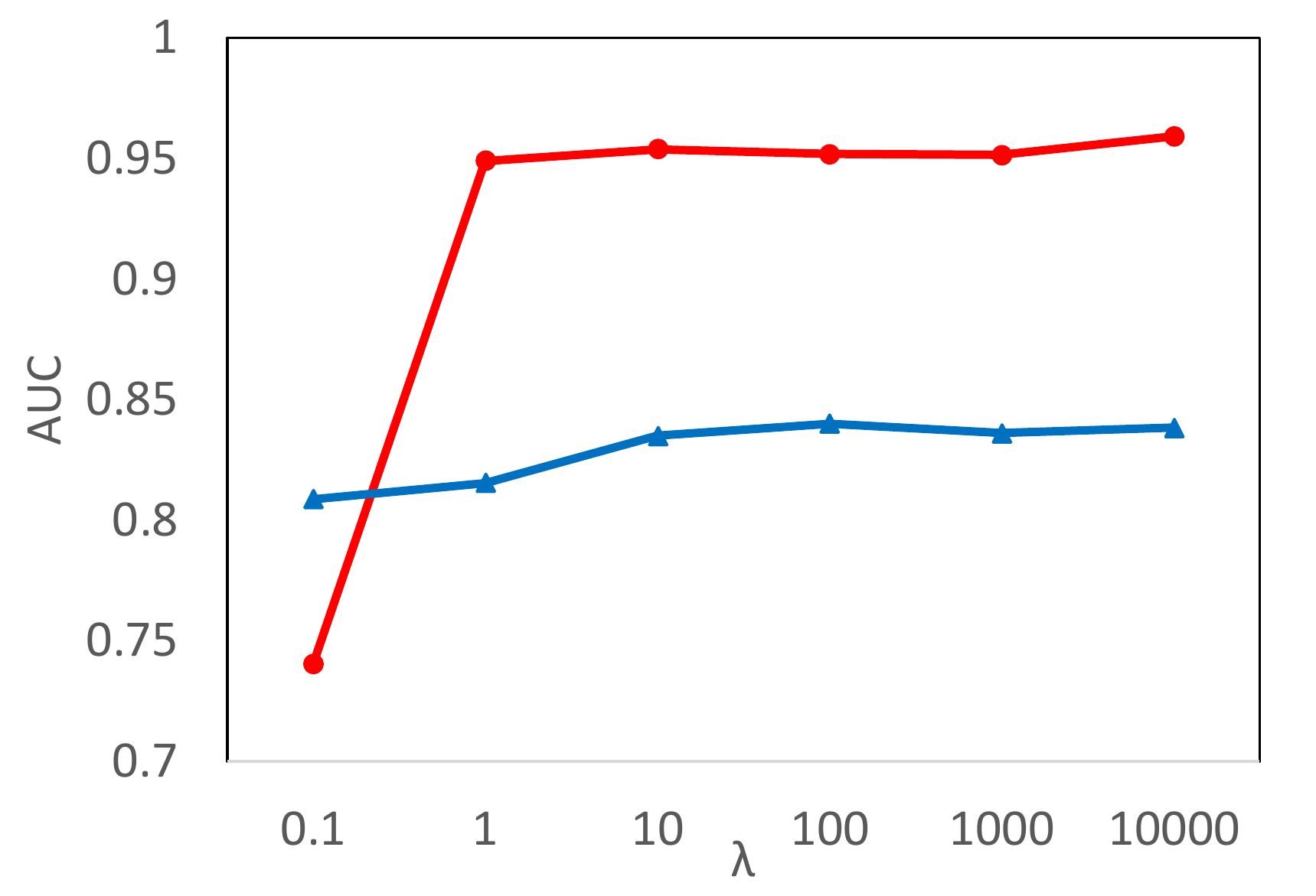}
      \subcaption{Photo}
  \end{minipage}
  \begin{minipage}{0.195\hsize}
      \centering
      \includegraphics[width=3.55cm]{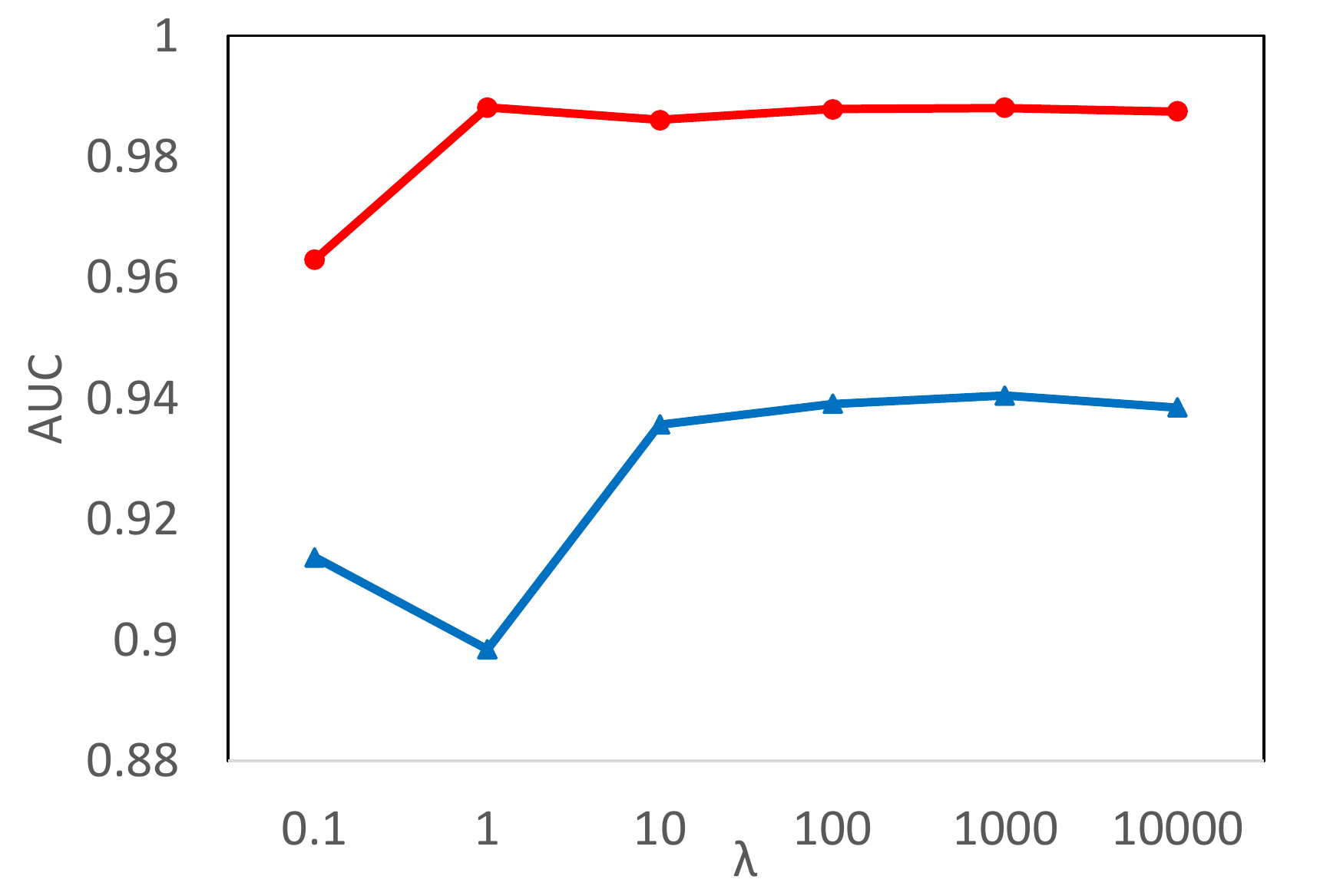}
      \subcaption{Comp}
  \end{minipage}
  \caption{Average test AUCs of each dataset when $\lambda$ was changed.}
  \label{fig_lam}
\end{figure*}
\begin{figure*}[t]
  \begin{minipage}{0.195\hsize}
      \centering
      \includegraphics[width=3.55cm]{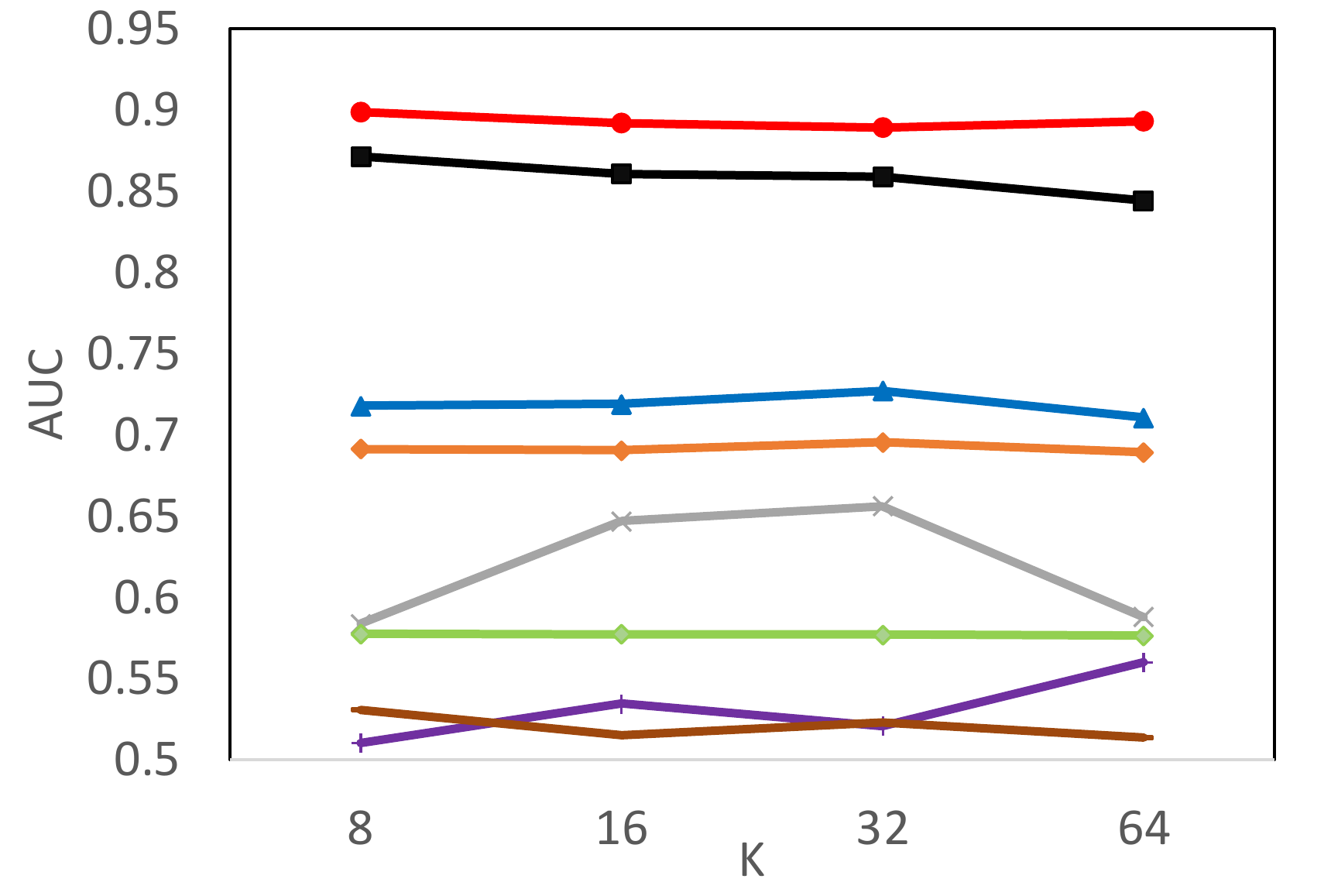}
      \subcaption{Cora}
  \end{minipage}
  \begin{minipage}{0.195\hsize}
      \centering
      \includegraphics[width=3.55cm]{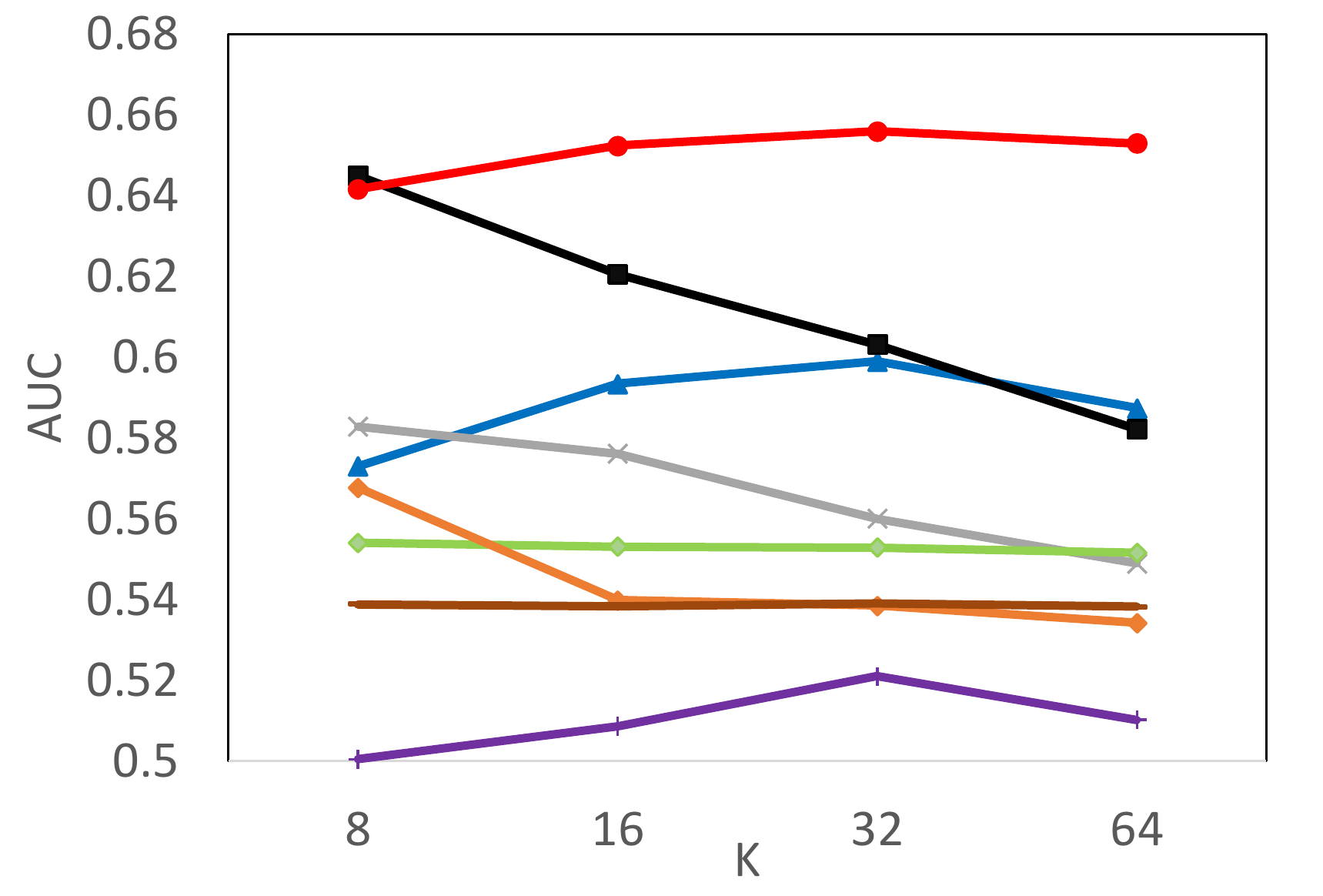}
      \subcaption{Cite}
  \end{minipage}
  \begin{minipage}{0.195\hsize}
      \centering
      \includegraphics[width=3.55cm]{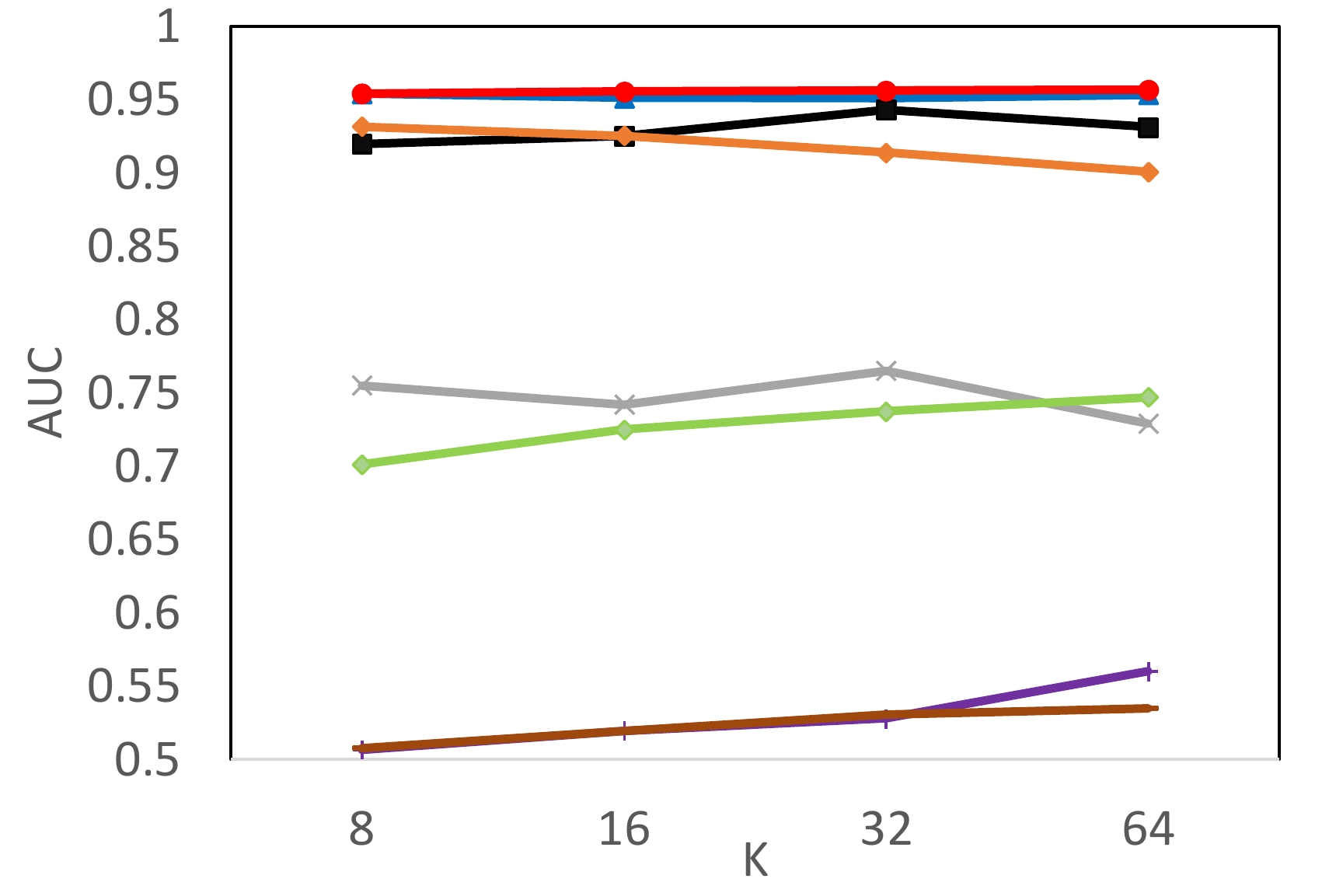}
      \subcaption{Pub}
  \end{minipage}
   \begin{minipage}{0.195\hsize}
      \centering
      \includegraphics[width=3.55cm]{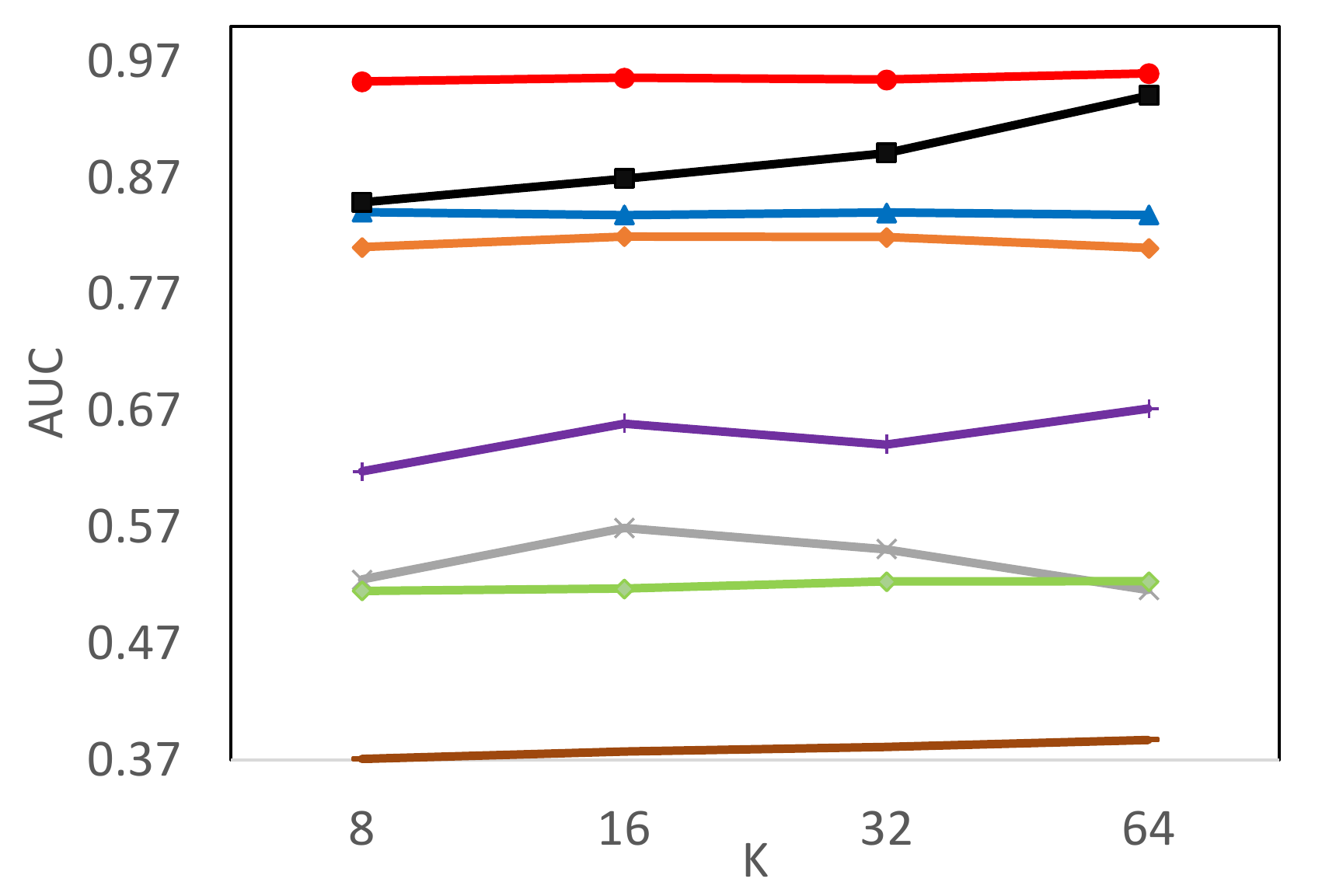}
      \subcaption{Photo}
  \end{minipage}
  \begin{minipage}{0.195\hsize}
      \centering
      \includegraphics[width=3.55cm]{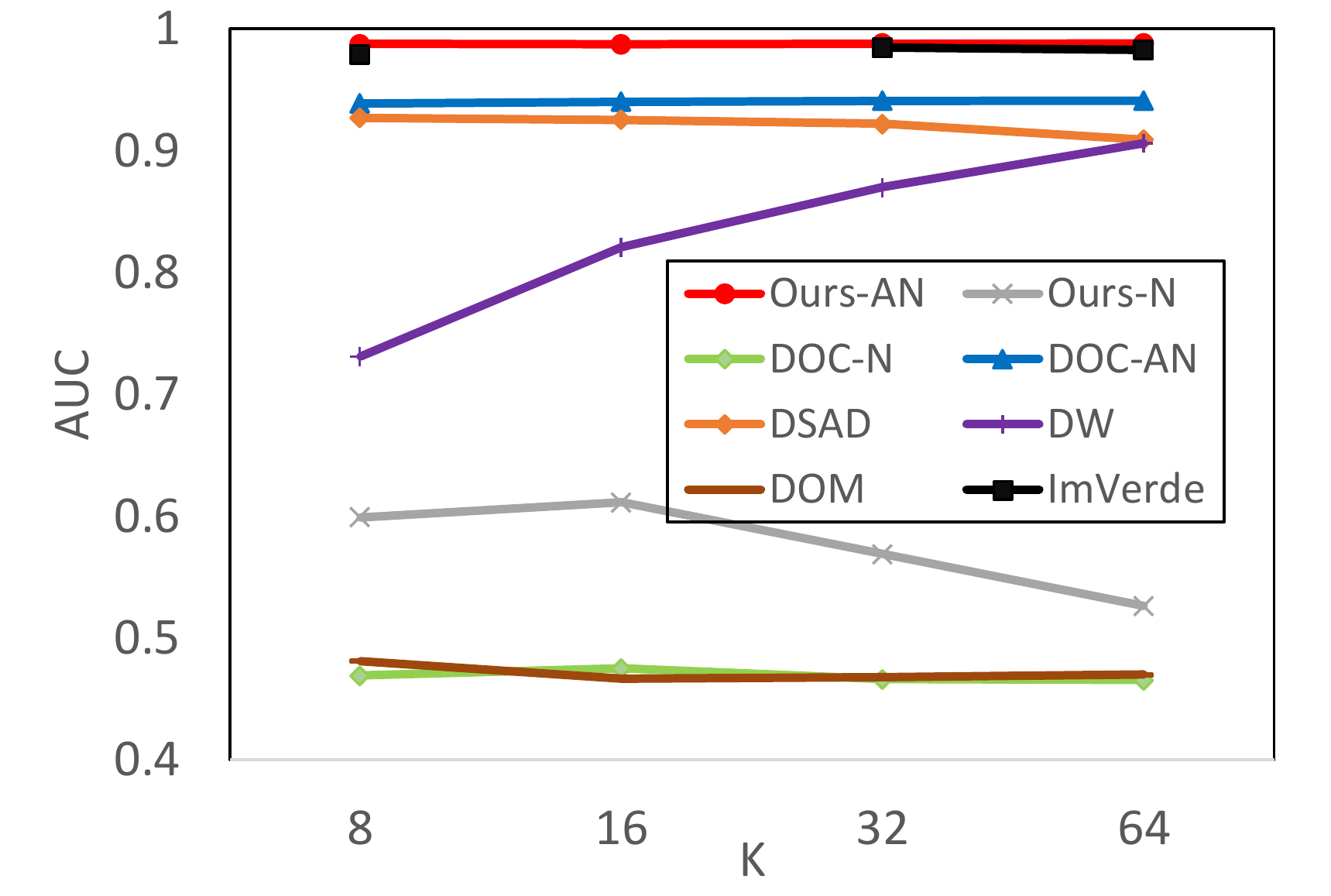}
      \subcaption{Comp}
  \end{minipage}
  \caption{Average test AUCs of each dataset when $K$ was changed.}
  \label{fig_k}
\end{figure*}

Third, we investigated the dependency of the regularization weight for the AUC regularizer $\lambda$ for Ours-AN and DOC-AN, which use the regularizer.
Figure \ref{fig_lam} shows the average test AUCs by changing $\lambda$ when 
the rate of labeled and all instances was 2.5\%.
For all datasets, Ours-AN outperformed DOC-AN in almost all $\lambda$, which indicates the robustness of the proposed method 
against $\lambda$.
The best $\lambda$ of Ours-AN differed across datasets. 
With Photo, large $\lambda$, which corresponds to minimize the AUC loss only in \eqref{final},
performed the best although the small value ($\lambda=1$) performed better with Cora, Cite, Pub, and Comp.

Fourth, we investigated the dependency of the dimension of embeddings $K$ for the proposed method.
Figure \ref{fig_k} shows the average test AUCs by changing $K$
when the rate of labeled and all instances was 2.5\%.
We compared Ours-AN and Ours-N with the embedding based methods: DOC-AN, DOC-N, DSAD, DW, DOM, and
ImVerde.
Ours-AN consistently performed well in all $K$ for all datasets.
As for methods that do not use anomalies for training, 
Ours-N performed better than DOC-N with almost all $K$'s.
Overall, these results suggest the proposed method is robust against the value of $K$. 

Lastly, we investigated the training times of 500 epochs for Ours-AN, Ours-N, and SLGCN on Cora when 
10\% of all instances were labeled and $K=32$.
The training times of Ours-AN, Ours-N, and SLGCN were 4.92, 3.56, and 2.46 seconds, respectively.
Since Ours-AN uses the AUC regularizer,
it took more training time than Ours-N.
However, the proposed method was able to learn fast enough. 

\section{Conclusions}
In this paper, we proposed a novel semi-supervised anomaly detection method on attribute graphs.
The proposed method utilizes graph GCNs to extract node embeddings considering both label information of small nodes and attribute information of all nodes with the graph structure.
To learn useful node embeddings for anomaly detection,
the proposed method minimizes the volume of the hypersphere that encloses normal node embeddings
while embedding anomalous ones outside the hypersphere.
In experiments using five real-world attributed graph datasets, the proposed method 
outperformed various existing anomaly detection methods in settings in which
both anomalous and normal labels or only normal labels are available for training.

\nocite{langley00}


\end{document}